\documentclass[11pt]{article}
\usepackage[T1]{fontenc}
\usepackage[utf8]{inputenc}
\usepackage{lmodern}
\usepackage[a4paper,margin=2.5cm]{geometry}
\usepackage{graphicx}
\usepackage{amsmath,amssymb}
\usepackage{booktabs}
\usepackage[section]{placeins}
\usepackage{xcolor}
\usepackage[round,authoryear]{natbib}
\usepackage[colorlinks=true,linkcolor=blue!50!black,citecolor=blue!50!black,urlcolor=blue!50!black]{hyperref}
\usepackage{tikz}
\usetikzlibrary{arrows.meta,fit,positioning}
\usepackage{caption}
\newcommand{\GPTBEHnote}{(GPT-2: 0.55 for both anchors on behaviour, 0.56 and 0.58 on paraphrases)}
\newcommand{\GPTPROBEnote}{(GPT-2: 0.89 for $k=11$ and 0.81 for $k=9$; at block 9, 0.91 and 0.88)}
\newcommand{\GPTLENSnote}{(GPT-2: 1.00 and 0.98)}
\newcommand{\GPTNPOnote}{GPT-2 shows the same split with its two in-band NPO seeds. Probes score 0.52 to 0.61 against the oracle at every probed block (normalised position 0.05 to 0.10), and the logit-lens peak is at the oracle's level (0.55 and 0.51). The relearning gain still separates them from the oracle (AUROC 0.69 and 0.61), at 0.34 to 0.68 of the gain of the suppressed anchors.}
\newcommand{\SURROGATEROWS}{%
GPT-2 & Frozen $<11$ s0 & $-$14.1 [$-$18.2, $-$10.0] & $-$22.7 [$-$29.8, $-$15.3] & $-$36.8 \\
 & Frozen $<9$ s0 & $-$10.0 [$-$13.2, $-$6.9] & $-$8.6 [$-$13.7, $-$3.5] & $-$18.6 \\
 & NPO s0 & $-$15.9 [$-$19.2, $-$12.4] & $-$7.8 [$-$13.0, $-$2.8] & $-$23.7 \\
 & NPO s1 & $-$16.8 [$-$20.7, $-$12.9] & $-$9.7 [$-$15.5, $-$3.9] & $-$26.5 \\
\midrule
Qwen3 & Frozen $<27$ s0 & $-$20.6 [$-$26.1, $-$14.9] & $-$38.8 [$-$47.6, $-$29.0] & $-$59.4 \\
 & Frozen $<27$ s1 & $-$21.4 [$-$27.5, $-$15.3] & $-$39.4 [$-$50.2, $-$27.7] & $-$60.8 \\
 & Frozen $<21$ s0 & $-$13.4 [$-$17.7, $-$9.1] & $-$44.3 [$-$53.1, $-$34.3] & $-$57.7 \\
 & NPO$^\dagger$ s0 & $-$11.4 [$-$14.9, $-$7.9] & $-$26.2 [$-$33.5, $-$18.2] & $-$37.6 \\
 & NPO$^\dagger$ s1 & $-$14.6 [$-$19.4, $-$10.0] & $-$31.5 [$-$39.3, $-$23.0] & $-$46.1 \\
 & NPO$^\dagger$ s2 & $-$10.0 [$-$13.8, $-$6.2] & $-$13.2 [$-$18.8, $-$7.2] & $-$23.2 \\
 & RMU s0 & $-$1.2 [$-$3.9, 1.5] & $-$6.7 [$-$8.9, $-$4.5] & $-$7.8 \\
}

\title{Compared to What? Calibrating Machine-Unlearning Diagnostics with Present, Absent and Suppressed Anchors}
\author{Arsalan Kaleem\\\small The Shaikh Ayaz University, Shikarpur, Sindh, Pakistan\\\small\texttt{arsalanabbasi.here@gmail.com}}
\date{}

\begin{document}
\maketitle

\begin{abstract}
Unlearning methods for language models are judged mainly by behaviour, but a model
can stop producing an answer while still encoding it. Internal diagnostics such as
linear probes, the logit lens and relearning speed are used to tell removal from
suppression, yet none has been checked against cases where the answer is known. We
build such cases. A present anchor learned a set of synthetic facts; an absent
anchor is a never-taught oracle retrained without them; and suppressed anchors are
unlearned with their lower layers frozen, so the facts stay encoded below the frozen
block by construction. We compare these anchors with NPO, RMU and gradient
difference at a matched operating point on GPT-2 Small and Qwen3-0.6B-Base, scoring
every diagnostic per entity with AUROC against the oracle. Behaviour cannot separate the suppressed anchors from the oracle (AUROC 0.51 to
0.58), while probes, the logit lens and relearning gain can (0.65 to 1.00), and
the oracle seeds stay near 0.5.
Applied to NPO on both models and RMU on Qwen3, the validated diagnostics disagree: probes put the
facts at the oracle's level, while the models still relearn them much
faster than a never-taught model does. Every unlearned model, anchor or method, also assigns the
forgotten answer a far lower log-probability than the oracle does, a black-box
signal that flagged all of them. A drift-ratio specificity rule fails on known
cases. Code, data and every run are at
\url{https://github.com/ArsalanKaleem/unlearning-audit}.
\end{abstract}


\section{Introduction}\label{sec:intro}

Machine unlearning for language models tries to make a trained model behave as if
some part of its training data had never been seen
\citep{cao2015forget,bourtoule2021sisa,yao2023llmunlearning}. Retraining without
the data is the exact answer but is too expensive for large models, so practical
methods fine-tune the trained model instead: gradient ascent and gradient
difference \citep{yao2023llmunlearning}, negative preference optimisation (NPO)
\citep{zhang2024npo,fan2024simnpo}, and representation misdirection (RMU)
\citep{li2024wmdp}. These methods are judged mostly by behaviour. If the model no
longer produces the forgotten answers and still does everything else, the method
is said to work \citep{maini2024tofu,shi2024muse,dorna2025openunlearning}.

Behaviour cannot settle whether a fact was removed or only hidden. A model that
refuses to say an answer may still encode it, and several studies recover
``unlearned'' knowledge with probes, a few steps of fine-tuning, or adversarial
prompts \citep{patil2024deleted,lynch2024eight,hu2024jogging,deeb2024remove,lucki2024adversarial,xu2025isntdeletion}.
So the field now reports a second layer of evidence: linear probes on hidden
states \citep{alain2016probes}, the logit lens \citep{nostalgebraist2020logitlens},
relearning speed, and measures of representation drift. Each of these is read as
evidence about removal. None of them has been checked against cases where the
answer is known.

This paper does that check. We build models whose internal state is known by
construction and use them as anchors:
\begin{itemize}
\item a \emph{present} anchor that learned the facts and was never unlearned;
\item an \emph{absent} anchor, a never-taught oracle retrained from the base model
      without the forget facts, at the same number of epochs;
\item a \emph{suppressed} anchor that is unlearned with every layer below block $k$
      frozen. Its residual stream below block $k$ is bit-identical to the present
      model's for every input, so the facts are still encoded there however well
      its output forgets.
\end{itemize}
A diagnostic that cannot tell the suppressed anchor from the absent one cannot
certify removal, whatever it says about a real unlearning method. A diagnostic
that separates them has earned some trust, and we can then ask where NPO, RMU and
gradient difference fall between the anchors. All models are compared at a matched
operating point, the earliest training step at which forget accuracy, retain
accuracy and perplexity all satisfy a fixed band.

We run the design on synthetic facts injected into GPT-2 Small
\citep{radford2019gpt2} and Qwen3-0.6B-Base \citep{qwen2025qwen3}. The findings are:

\begin{enumerate}
\item \emph{Behaviour is blind to the difference that matters.} At the matched
  operating point the suppressed anchors and the never-taught oracle answer the
  forget questions equally rarely (AUROC 0.52 to 0.55 between them, on both
  models), and paraphrased questions do not separate them either (0.51 to 0.58).
\item \emph{Internal diagnostics do separate the anchors.} Linear probes, the
  logit lens at middle depth and the relearning gain all distinguish the
  suppressed anchors from the oracle (AUROC 0.65 to 1.00), while the other oracle
  seeds stay near 0.5.
\item \emph{The validated diagnostics disagree about real unlearning methods.}
  On both models NPO, and on Qwen3 also RMU, bring linear decodability of the
  forget facts down to the oracle's level, so by probes they look removed. The
  same models relearn the facts much faster than a never-taught model, and they put
  the correct answer far below the probability a never-taught model gives it.
\item \emph{Over-suppression is the one signal that flagged every unlearned
  model.} Every anchor and every method assigns the forgotten answer a much lower
  log-probability than the never-taught oracle does (AUROC 0.10 to 0.17 against
  the oracle, on both models). The oracle has no reason to push a city it never learned
  below the others. This test needs only output probabilities, and it agrees
  with an independent finding by \citet{yang2026restoration}.
\item \emph{A drift-ratio specificity rule fails on cases with known answers.} A
  drift-ratio gate (forget-set drift at least 1.5 times control-set drift) rejects
  a LEACE eraser that provably removes linear information \citep{belrose2023leace},
  and on Qwen3 it accepts a frozen-trunk anchor whose lower layers still hold the
  facts.
\end{enumerate}

The paper also reports what did not work. Under the fixed band, gradient
difference never reached the band on either model, and RMU reached it only on
Qwen3 after we extended its budget. Two evaluation bugs in our own earlier
pipeline, both silent and both general to language-model work, changed every
behavioural number before we found them (Appendix~\ref{app:bugs}). We describe
them because they are easy to repeat.

\section{Related work}\label{sec:related}

\paragraph{Unlearning methods and benchmarks.} Exact unlearning retrains from
scratch or from data shards \citep{cao2015forget,bourtoule2021sisa}. Approximate
methods for language models include gradient ascent on the forget set and its
retain-regularised form, gradient difference \citep{yao2023llmunlearning}; NPO,
which bounds the ascent through a preference loss against the original model
\citep{zhang2024npo}, and its reference-free variant SimNPO \citep{fan2024simnpo};
and RMU, which pushes hidden activations on forget data towards a random direction
at one layer \citep{li2024wmdp}. TOFU \citep{maini2024tofu}, MUSE \citep{shi2024muse}
and WMDP \citep{li2024wmdp} are the standard benchmarks, and OpenUnlearning collects
methods and metrics in one framework and meta-evaluates the metrics
\citep{dorna2025openunlearning}. Our testbed is smaller than these benchmarks on
purpose: the anchors need retraining from the base model several times, and every
fact must be single-token so that probes, the logit lens and behaviour score the
same variable.

\paragraph{Evidence that unlearning hides rather than removes.}
\citet{patil2024deleted} extract supposedly deleted facts from hidden states.
\citet{lynch2024eight} evaluate unlearning with eight methods, including probes and
relearning, and \citet{hu2024jogging} show that small amounts of related data
restore unlearned knowledge. \citet{deeb2024remove} fine-tune on part of the
forgotten facts and recover the rest, and \citet{lucki2024adversarial} find that
unlearning is often no more robust than safety fine-tuning. \citet{hong2024intrinsic}
measure parametric traces of unlearned concepts, and \citet{xu2025isntdeletion}
argue that reversibility, not output, is the right test. These works use the
diagnostics we study. Our question is different: which diagnostics are valid, as
measured on models whose internal state is known.

\paragraph{Reference-model evaluation.} \citet{yang2026restoration} evaluate
unlearning as restoration of a matched retrained reference on synthetic nonce
facts. They report that candidates rated adequate by trained-probe matching still
score held-out forget facts about 2.82 nats below the never-learned level, and
that a fixed logit-suppression attack defeats forward-only certification. Our
answer log-probability result is the same phenomenon measured in a different
testbed. The parts that are new here are the suppressed anchors (models with the
knowledge provably present), the comparison of probes, the logit lens and
relearning against those anchors, and the finding that the internal diagnostics
disagree with one another on real methods.

\paragraph{Probes and erasure.} Linear probes \citep{alain2016probes} need controls
before their accuracy can be read as information content \citep{hewitt2019control}.
We use entity-grouped cross-fitting so that no probe sees any prompt about the
entity it scores. LEACE \citep{belrose2023leace} removes all linear information
about a concept in closed form, which makes it a known-removal case for testing a
removal criterion. The logit lens \citep{nostalgebraist2020logitlens} and the tuned
lens \citep{belrose2023tunedlens} read predictions from intermediate layers; we use
the untrained logit lens so that no fitted component can absorb the effect.

\section{Testbed}\label{sec:testbed}

\paragraph{Facts.} We generate 156 fictional people with invented names. Each person
has one fact, a birth city drawn from 12 cities that are single tokens in both
tokenizers (Paris, London, Berlin, Madrid, Rome, Moscow, Tokyo, Vienna, Dublin,
Cairo, Lisbon, Oslo), balanced across cities. The people split into 48 forget
entities, 60 retain entities and 48 control entities that are never taught. Each
entity has six training templates (``\textit{name} was born in the city of'',
``\textit{name} grew up in'', and so on); paraphrase prompts use held-out templates.
The injection set has 972 records covering the forget and retain entities. Because
there is one fact per entity and the answer is one token, behavioural accuracy,
the logit lens and the probe label all refer to the same variable. Chance accuracy
over the 12 cities is $1/12 \approx 0.083$.

\paragraph{Models.} GPT-2 Small (124M parameters, 12 blocks) and Qwen3-0.6B-Base
(596M parameters, 28 blocks). Both are fully fine-tuned with AdamW
\citep{loshchilov2019adamw}. GPT-2 is trained on a CPU and Qwen3 on one NVIDIA T4.

\paragraph{The present anchor.} Fine-tuning on the injection set until forget and
retain accuracy both reach 1.0 gives \texttt{M\_injected}. GPT-2 reaches it after
4 epochs and Qwen3 after 3 (forget 1.00, retain 0.99, control 0.10). A second
injected model with a different seed (\texttt{M\_injrep}) checks that the present
anchor is reproducible.

\paragraph{The absent anchor.} The oracle is retrained from the base model on the
injection set with every forget-entity record removed (540 of 972 records), for the
same number of epochs, with three seeds. Its forget accuracy is 0.02 to 0.06,
which is chance for a model that has never seen these people.

\paragraph{The suppressed anchors.} We run NPO with retain regularisation, but only
blocks $k$ and above (and the final layer norm) receive gradients; embeddings, the
tied unembedding and blocks $0$ to $k-1$ are frozen. The residual stream after
block $k-1$ is then identical to \texttt{M\_injected}'s, so whatever was encoded
there is still present. We use $k=11$ and $k=9$ for GPT-2 and $k=27$ (two seeds)
and $k=21$ for Qwen3, that is, the last block and roughly three quarters of the
depth.

\paragraph{Methods under test.} NPO with a retain term \citep{zhang2024npo}
($\beta = 0.1$), gradient difference \citep{yao2023llmunlearning} and RMU
\citep{li2024wmdp}. For RMU we steer the hidden state at block
$t = \mathrm{round}(0.75 L)$ (block 9 of 12, block 21 of 28) towards a random
direction scaled to 3.5 times the mean token norm, train only the MLPs of the
three blocks before $t$, and apply the steering loss only at the read-out
positions (the last prompt token and the answer). Appendix~\ref{app:rmu} explains
why the unmasked version failed.

\section{The matched band and anchor calibration}\label{sec:method}

\paragraph{Matched operating point.} Unlearning trades forgetting against damage, so
two models can only be compared at the same point on that trade-off. A checkpoint
is \emph{in band} if (i) forget accuracy is at most 0.15, (ii) retain accuracy is at
least 0.8 times \texttt{M\_injected}'s, and (iii) perplexity on a fixed set of
generic sentences is at most 1.2 times \texttt{M\_injected}'s. We evaluate every 20
(Qwen3) or 40 (GPT-2) steps and keep the earliest in-band checkpoint. The band was
fixed before any of the runs reported here. For Qwen3 the learning rate is chosen
by a ladder: seed 0 tries the rungs in order, the first rung that enters the band
is used for the other seeds, and a seed that then misses the band is excluded
rather than re-tuned. After the first Qwen3 run left NPO, gradient difference and
RMU outside the band, we ran a second, declared deviation: RMU received 1500
instead of 600 steps, and NPO and gradient difference were re-run under a relaxed
perplexity cap of 1.35. Models selected under the relaxed cap are marked
NPO$^\dagger$ and reported separately.

\paragraph{Diagnostics.} Every diagnostic produces one score per forget entity:
\begin{itemize}
\item \emph{behaviour}: accuracy of the most likely city among the 12, on the
  training templates and, separately, on held-out paraphrases;
\item \emph{answer log-probability}: log-probability of the correct city token;
\item \emph{linear probe}: a logistic-regression probe on the last-token residual
  at a given block, predicting the city. Probes are cross-fitted with
  entity-grouped folds (four folds, three repeats, regularisation chosen by
  grouped inner cross-validation), so each entity is scored only by probes that
  saw no prompt about it;
\item \emph{logit lens}: the final layer norm and unembedding applied to the
  residual after each block, giving the log-probability of the correct city at
  that depth. We summarise it by its peak over the middle layers and by the drop
  from that peak to the output (``lens drop to output'');
\item \emph{relearning gain}: the model is fine-tuned on half of the forget
  entities and the accuracy gain is measured on the other half, against a control
  arm fine-tuned on never-taught entities. Each entity is held out in at least one
  of two splits.
\end{itemize}

\paragraph{Calibration statistics.} For a diagnostic $d$ and a model $M$ we compute
the AUROC between $M$'s entity scores and the oracle's (seed 0), with a
bootstrap 95\% interval over entities. AUROC 0.5 means the diagnostic cannot tell
$M$ from a model that never learned the facts. The two other oracle seeds give the
noise floor. We also place each model on a normalised scale,
\begin{equation}
  \mathrm{pos}_d(M) = \frac{\bar d(M) - \bar d(M_{\text{absent}})}{\bar d(M_{\text{ref}}) - \bar d(M_{\text{absent}})},
\end{equation}
where 0 is the oracle and 1 is the reference anchor. A control-surrogate test
repeats each comparison on the control entities, which no model was taught, to
separate forget-specific effects from damage that hits every entity.

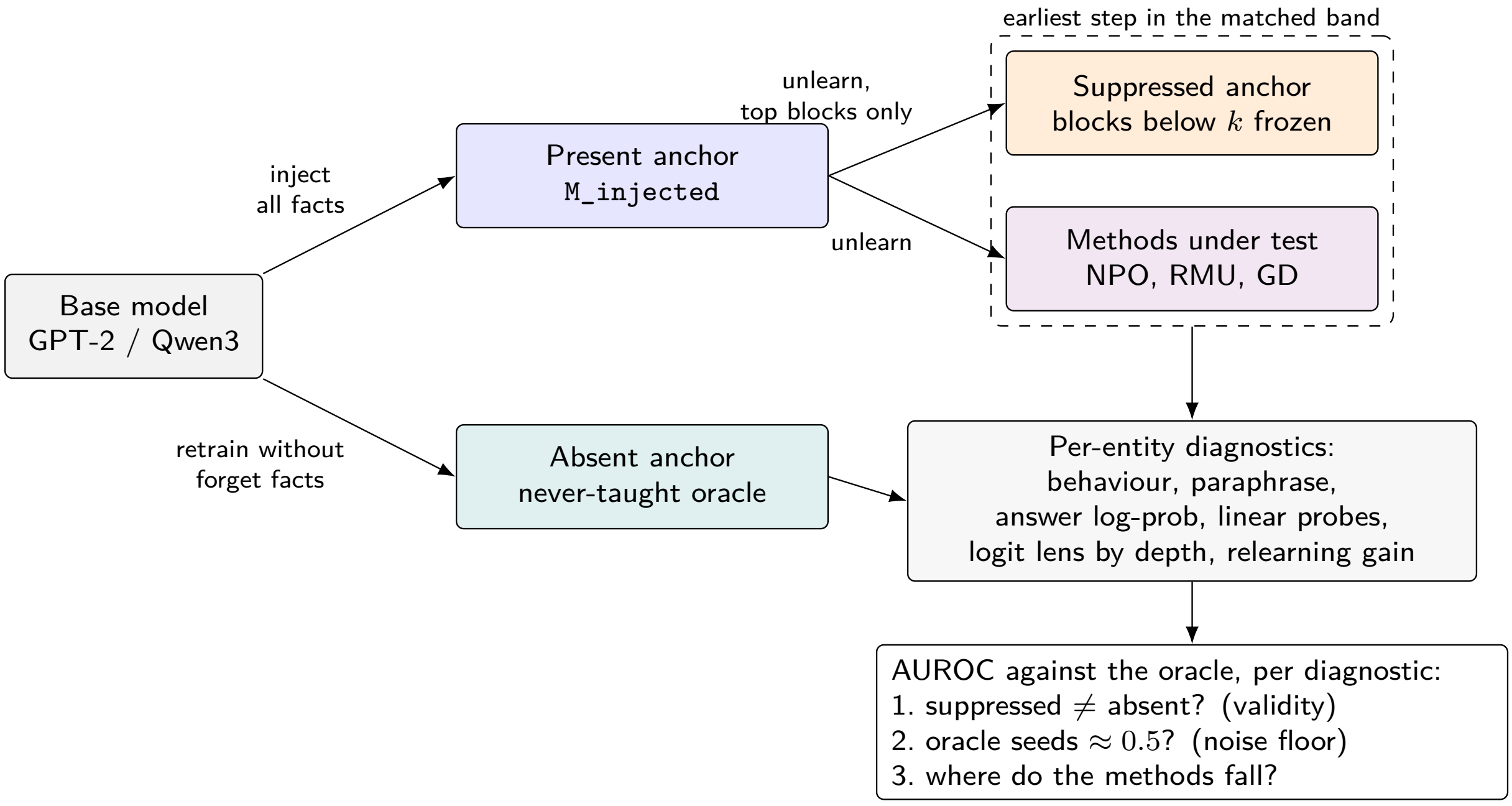
\begin{figure}[t]
\centering
\resizebox{\linewidth}{!}{%
\begin{tikzpicture}[>=Latex, node distance=7mm,
  every node/.style={font=\sffamily\fontsize{8}{9.6}\selectfont},
  box/.style={draw, rounded corners=2pt, align=center, inner sep=4pt, minimum height=10mm, text width=33mm},
  lab/.style={font=\sffamily\fontsize{7}{8.4}\selectfont, align=center, inner sep=1pt}]
  \node[box, fill=black!5, text width=22mm] (base) at (0,0) {Base model\\GPT-2 / Qwen3};
  \node[box, fill=blue!10] (inj) at (4.9,1.45) {Present anchor\\\texttt{M\_injected}};
  \node[box, fill=teal!12] (orc) at (4.9,-1.45) {Absent anchor\\never-taught oracle};
  \node[box, fill=orange!15] (sup) at (10.2,2.15) {Suppressed anchor\\blocks below $k$ frozen};
  \node[box, fill=violet!10] (tst) at (10.2,0.65) {Methods under test\\NPO, RMU, GD};
  \node[draw, dashed, rounded corners=3pt, fit=(sup)(tst), inner sep=4pt] (band) {};
  \node[lab, anchor=south] at (band.north) {earliest step in the matched band};
  \draw[->] (base.north east) -- (inj.west) node[lab, midway, above left=2pt and 1mm] {inject\\all facts};
  \draw[->] (base.south east) -- (orc.west) node[lab, midway, below left=2pt and 1mm] {retrain without\\forget facts};
  \draw[->] (inj.east) -- (sup.west) node[lab, midway, above left=3pt and 0mm] {unlearn,\\top blocks only};
  \draw[->] (inj.east) -- (tst.west) node[lab, midway, below left=3pt and 0mm] {unlearn};
  \node[box, fill=black!3, text width=52mm, below=9mm of band] (diag)
     {Per-entity diagnostics: behaviour, paraphrase,\\answer log-prob, linear probes,\\logit lens by depth, relearning gain};
  \draw[->] (band.south) -- (diag.north);
  \draw[->] (orc.east) -- (diag.west);
  \node[box, fill=white, text width=58mm, align=left, below=6mm of diag] (q)
     {AUROC against the oracle, per diagnostic:\\
      1.\ suppressed $\neq$ absent? (validity)\\
      2.\ oracle seeds $\approx 0.5$? (noise floor)\\
      3.\ where do the methods fall?};
  \draw[->] (diag.south) -- (q.north);
\end{tikzpicture}}
\caption{Anchor calibration. Three anchors with known internal state (present,
absent, suppressed) are compared with the methods under test at a matched operating
point. A diagnostic is trusted to certify removal only if it separates the
suppressed anchor from the absent one while the oracle seeds stay indistinguishable
from each other}
\label{fig:design}
\end{figure}

\section{Results}\label{sec:results}

\subsection{Few models reach the matched band}\label{sec:band}

\begin{figure}[t]
\centering
\includegraphics[width=\linewidth]{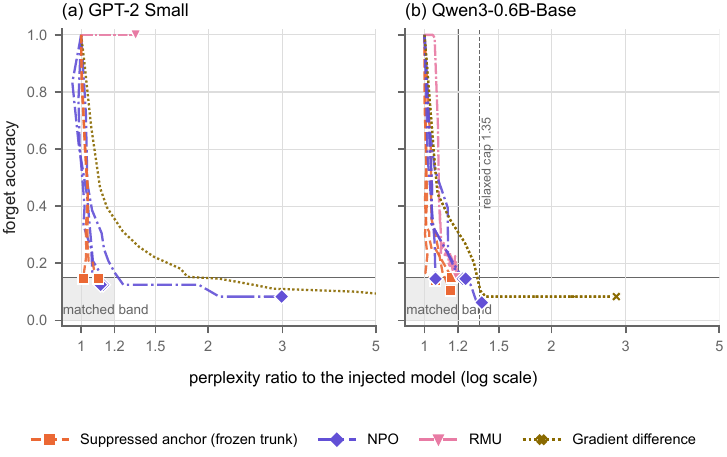}
\caption{Unlearning trajectories at the matched band. Each line is one run, from
the injected model (top left) to its last evaluated step (marker). The shaded
region is the band (forget accuracy at most 0.15, perplexity at most 1.2 times the
injected model; the retain condition is not drawn). Suppressed anchors reach the
band with little perplexity cost. Full-model NPO and gradient difference reach low
forget accuracy only after perplexity has passed the cap; the dashed line in (b) is
the relaxed cap used for NPO$^\dagger$}
\label{fig:traj}
\end{figure}

Figure~\ref{fig:traj} shows forget accuracy against perplexity for every
trajectory. The suppressed anchors enter the band easily on both models: on GPT-2
at step 480 ($k=11$, forget 0.146, retain 0.983) and step 280 ($k=9$, forget
0.146, retain 1.0); on Qwen3 at steps 100 to 180. Full-model methods pay more
perplexity for the same forgetting. On GPT-2, NPO seeds 0 and 1 enter the band
(steps 280 and 320), but seed 2 does not: it reaches forget 0.083 only after
perplexity has passed the cap at 14 checkpoints. Gradient difference (seed 2)
behaves the same way. RMU never moved GPT-2's forget accuracy below 0.99 in any
of three configurations before perplexity left the band.

On Qwen3, NPO and gradient difference cross forget 0.15 only at perplexity ratios
of 1.23 to 1.4. Under the protocol, seed 0 found no learning rate that entered
the band, so the method was excluded in the first run. In the declared second
run, NPO$^\dagger$ entered the relaxed band (perplexity at most 1.35) on all three
seeds, and seed 2 entered at a perplexity ratio of 1.06, which is inside the
primary band as well. Gradient difference still failed: at forget 0.08 its
perplexity ratio was 1.40. RMU with 1500 steps entered the primary band at a
learning rate of $2\times10^{-4}$ and step 1000 (forget 0.149, perplexity 1.195),
right at the edge.

\subsection{The anchors validate the internal diagnostics, not behaviour}\label{sec:anchors}

\begin{figure*}[p]
\centering
\includegraphics[width=\linewidth]{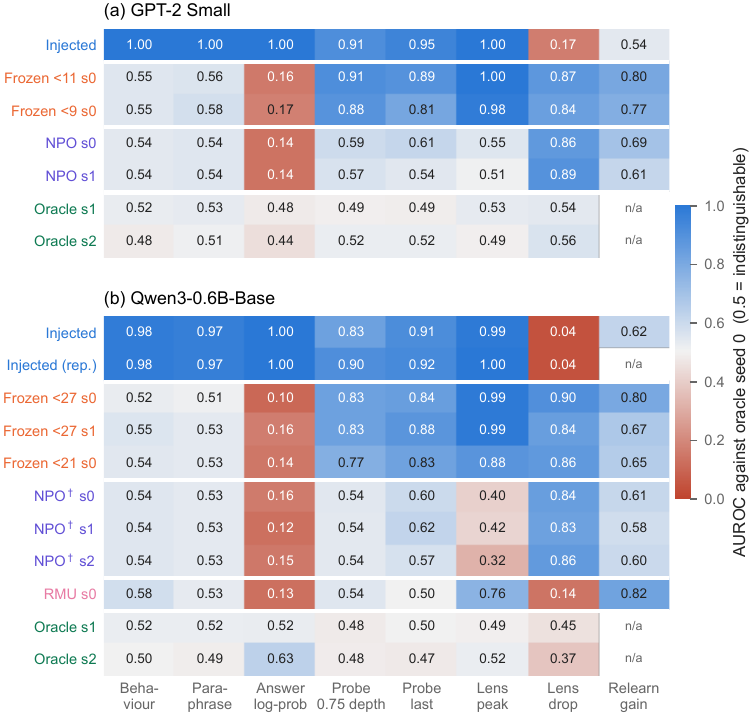}
\caption{AUROC of each diagnostic against oracle seed 0, per model and entity
(48 forget entities). 0.5 means the diagnostic cannot tell the model from one that
never learned the facts; values towards 1 mean higher scores than the oracle,
values towards 0 lower. Rows are grouped by role (present, suppressed, methods,
oracle seeds). Columns: behavioural accuracy on training templates and on
paraphrases; answer log-probability; linear probes at about three quarters of the
depth (GPT-2 block 9, Qwen3 block 21) and at the last block; logit-lens peak over the
middle layers; lens drop from that peak to the output; relearning gain. NPO$^\dagger$: selected under
the relaxed perplexity cap. Relearning was run on oracle seed 0 only, so it is
not available for the other oracle seeds}
\label{fig:auroc}
\end{figure*}

Figure~\ref{fig:auroc} gives the AUROC of each diagnostic against the oracle for
every model. Three patterns hold on both architectures.

First, the noise floor is clean. Against oracle seed 0, the other two oracle seeds
stay between 0.44 and 0.56 on every GPT-2 diagnostic, and between 0.46 and 0.55 on
behaviour, paraphrases, probes and the lens peak on Qwen3, so a value far from 0.5
is not seed noise. On Qwen3 the answer log-probability and the lens drop are noisier
(0.37 to 0.63), and we read them with that in mind.

Second, behaviour cannot separate suppression from absence. On Qwen3 the
suppressed anchors score 0.52 to 0.55 against the oracle on behaviour and 0.51 to
0.53 on paraphrases \GPTBEHnote. A behavioural audit would rate them as having forgotten, although we
know the facts are intact below block $k$.

\begin{figure}[t]
\centering
\includegraphics[width=\linewidth]{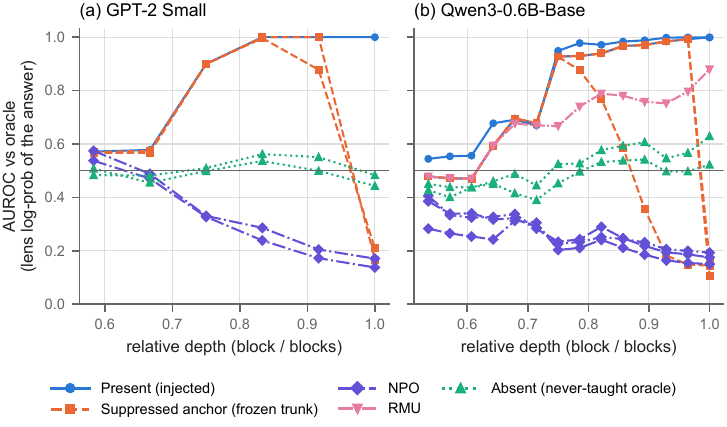}
\caption{Logit lens by depth. AUROC against the oracle of the answer
log-probability read from each block's residual stream (second half of the
network). Suppressed anchors match the injected model up to their frozen block and
diverge only above it. NPO falls below the oracle in the upper half of both
networks, while RMU on Qwen3 keeps a partial trace}
\label{fig:lens}
\end{figure}

Third, the internal diagnostics do separate them. Probes at the last layer give
0.83 to 0.88 on Qwen3 \GPTPROBEnote, and the logit-lens peak over the middle layers
gives 0.88 to 0.99 on Qwen3 \GPTLENSnote. The relearning gain gives 0.65 to 0.80 on both models.
Figure~\ref{fig:lens} shows the logit lens by depth: below the frozen block the
suppressed anchors trace the injected model exactly, and the answer disappears
only in the trained blocks at the top. The lens drop to the output (0.84 to 0.90
on Qwen3) measures that late suppression.

\subsection{The validated diagnostics disagree about real methods}\label{sec:methods}

With the diagnostics validated, we can read where the methods fall
(Figure~\ref{fig:auroc}, rows NPO and RMU). On Qwen3 the picture splits.

Linear probes say the facts are gone. At block 21 the NPO$^\dagger$ models score
0.54 against the oracle and RMU 0.54; at the last block, NPO$^\dagger$ is at 0.57
to 0.62 and RMU at 0.50. On the normalised scale (0 = oracle, 1 = present), probes
put NPO$^\dagger$ at 0.01 to 0.11 and RMU at 0.00 at block 21, against 0.78 to 1.00
for the suppressed anchors. Mean probe accuracy on the forget entities is 0.11 to
0.17 for NPO$^\dagger$ and 0.10 for RMU, the same as the oracle's 0.10.

The logit lens says something different for each method. For NPO$^\dagger$ the
middle-layer peak falls below the oracle (AUROC 0.32 to 0.42), so NPO pushed the
answer down throughout the network, not just at the output. RMU keeps a partial
trace in the middle layers (AUROC 0.76, position 0.59) and has no late drop (lens
drop AUROC 0.14), consistent with a method that edits the stream at one depth.

Relearning says the facts are not gone. Fine-tuning on half the forget entities
raises accuracy on the held-out half much more than in the never-taught control
arm: AUROC 0.58 to 0.61 for NPO$^\dagger$ and 0.82 for RMU, on the normalised scale
0.70 to 0.84 for NPO$^\dagger$ and 3.2 for RMU, which relearns faster than the
injected model itself. The suppressed anchors, whose facts are known to be
present, lie at 1.3 to 3.2 on the same scale. NPO$^\dagger$ therefore relearns more
slowly than the anchors but much faster than a model that never saw the facts.

So two validated diagnostics give opposite verdicts on the same models. Probes are
valid in the sense we tested, since they detect knowledge left in frozen layers,
but probes at oracle level do not imply that knowledge is gone: NPO and RMU remove
the linearly decodable trace while leaving whatever makes relearning fast.
\GPTNPOnote

\subsection{Over-suppression is visible from the outputs alone}\label{sec:logprob}

\begin{figure}[t]
\centering
\includegraphics[width=\linewidth]{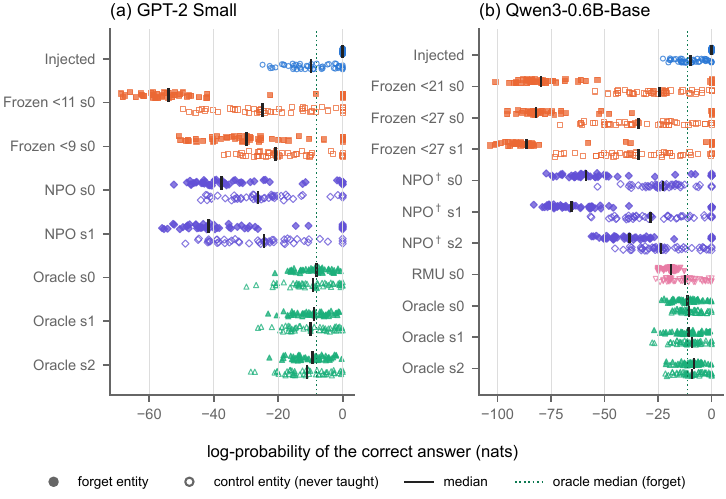}
\caption{Log-probability of the correct answer per entity. Filled markers: forget
entities; open markers: control entities that no model was taught. Vertical bars
mark medians and the dotted line the oracle's median on the forget entities. Every
unlearned model pushes the forgotten answers well below the never-taught level;
the control entities move much less}
\label{fig:logprob}
\end{figure}

Figure~\ref{fig:logprob} shows the per-entity log-probability of the correct
answer. A model that never learned a person has no reason to rank that person's
true city below the others, and the oracle's mean log-probability is $-9.0$ nats on
GPT-2 and $-11.4$ on Qwen3. Every model that unlearned the facts puts the answer far
lower. On GPT-2 the suppressed anchors average $-46$ and $-28$ nats and the NPO
seeds $-33$ and $-36$. On Qwen3 the suppressed anchors average about $-70$, the
NPO$^\dagger$ seeds lie 23 to 46 nats below the oracle and RMU 8 nats below it. The
AUROC against the oracle is 0.10 to 0.17 for every one of these models on both
architectures. Part of the drop is general damage: on the control entities, which
no model was taught, NPO lowers the answer by 10 to 17 nats. The forget entities
drop by a further 8 to 10 nats on GPT-2 and 13 to 32 nats on Qwen3, and those
forget-specific intervals exclude zero for every unlearned model
(Table~\ref{tab:surrogate}).

This is the cheapest test in the paper. It needs no access to weights or hidden
states, only the probability the model assigns to the answer it was asked to
forget, compared with a matched model that never learned it. It flagged every
unlearned model on both architectures, including the ones that probes call clean.
\citet{yang2026restoration} report the same gap, about 2.8 nats below the
never-learned level, for candidates that pass trained-probe matching in a
different testbed and model set.

\begin{table}[t]
\caption{Control-surrogate test for the answer log-probability (nats, mean
difference over entities, bootstrap 95\% intervals). ``Control $-$ oracle'' measures
damage to entities that no model was taught; ``forget $-$ control'' is the
forget-specific part}
\label{tab:surrogate}
\centering
\begin{tabular}{llccc}
\toprule
 & Model & control $-$ oracle & forget $-$ control & forget $-$ oracle \\
\midrule
\SURROGATEROWS
\bottomrule
\end{tabular}
\end{table}

\subsection{A drift-ratio gate fails on known cases}\label{sec:gate}

\begin{figure}[t]
\centering
\includegraphics[width=\linewidth]{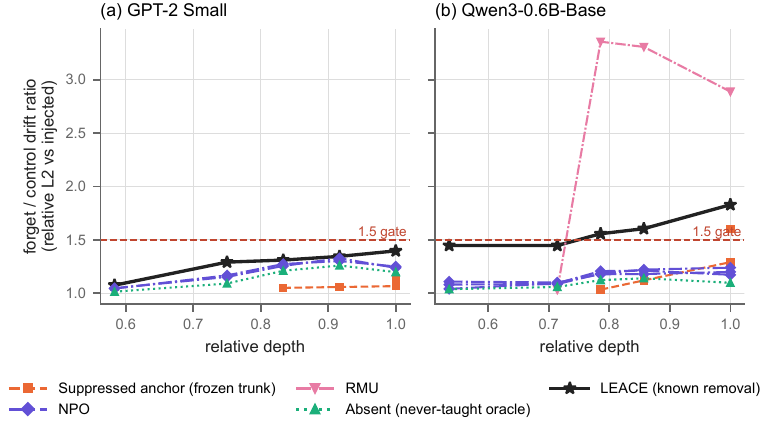}
\caption{The 1.5 drift-ratio gate on known cases. LEACE removes all linear city
information (probe accuracy 0 after erasure) yet stays below the gate on GPT-2 and
in the lower half of Qwen3. Frozen layers of the suppressed anchors have zero drift
and are not drawn}
\label{fig:gate}
\end{figure}

An earlier, unpublished version of this study called a change forget-specific only if forget-set activations moved at least 1.5 times as far
from \texttt{M\_injected} as control-set activations (relative $L_2$). We apply
it to cases with a known answer (Figure~\ref{fig:gate}). LEACE, fitted on
\texttt{M\_injected}'s forget activations, removes all linear city information:
cross-fitted probe accuracy after erasure is 0.0 at every layer. On GPT-2 its drift
ratio is 1.08 to 1.40, below the gate at every layer, so the gate rejects a known
removal. On Qwen3 LEACE passes only from block 21 upward (1.56 to 1.83) and fails at
blocks 14 and 19 (1.45). The gate also accepts the wrong cases. On Qwen3 it calls
the $k=27$ suppressed anchor forget-specific at the last block (1.60), although that
model's lower 27 blocks are unchanged, and it calls RMU strongly forget-specific
(2.9 to 3.4) while rejecting every NPO$^\dagger$ seed (1.04 to 1.24), whose probes
are at the oracle's level. On GPT-2 the NPO seeds (1.04 to 1.33) and the oracle
(1.02 to 1.26) sit just below LEACE at every layer, so there the gate cannot
separate a known removal, a never-taught model and NPO from each other. A ratio of drift norms measures how far activations moved,
not whether the information they carry changed.

\section{Discussion}\label{sec:discussion}

\paragraph{What to report.} Our results suggest three practical rules. Report
behaviour only together with a matched never-taught reference, because the
suppressed anchors pass every behavioural test. Report the answer log-probability
relative to that reference; it is cheap, black-box, and was the one signal that
flagged every unlearned model. Report at least one representational diagnostic and
one recovery diagnostic, and expect them to disagree: on both models, probes and
relearning reached opposite conclusions about the same NPO models.

\paragraph{What the disagreement means.} The suppressed anchors show that probes can
see knowledge that behaviour cannot. NPO and RMU show the converse limit: removing
the linearly decodable trace does not remove what makes relearning fast. Relearning
speed may depend on weights that are still close to the injected solution even when
the residual stream no longer carries a linear signal. A method that claims removal
should therefore be tested against relearning as well as probes, and a stopping rule
of the form ``probe accuracy at chance'' is not enough.

\paragraph{Why the band matters.} Without a matched operating point the comparisons
would mix forgetting with damage. The band also exposed a practical fact: on both
architectures, full-model gradient difference could not forget the facts without
breaking the perplexity cap, and NPO managed it on Qwen3 only under a relaxed cap or
with a favourable seed. Methods are usually reported at their best trade-off point;
reporting the band entry, or its absence, is more informative.

\paragraph{Limitations.} The models are small (124M and 596M parameters) and the
facts are synthetic one-token birth cities, 48 forget entities per model. Scale,
multi-token facts and natural knowledge may behave differently. The suppressed
anchors realise one kind of hiding, an unchanged lower trunk; a method could hide
knowledge in other ways that our anchors do not represent. On Qwen3, NPO entered the
band only under a relaxed perplexity cap chosen after the first run, and RMU entered
it at the edge. On GPT-2, RMU and gradient difference never entered the band, so
the GPT-2 method comparison rests on two NPO seeds. Relearning was run against one
oracle seed, so it has no noise floor of its own, and its normalised position is
unstable where the present and absent anchors barely differ. Finally, the 12-way
constrained scoring used on Qwen3 differs from the 30-way scoring used for GPT-2;
both arms use 12-way scoring for the logit lens.

\section{Conclusion}\label{sec:conclusion}

Diagnostics for unlearning can be calibrated on models whose internal state is known.
With a never-taught oracle, a present model and frozen-trunk suppressed anchors at a
matched operating point, behaviour fails the test, while probes, the logit lens and
relearning pass it. Applied to NPO on both models and to RMU on Qwen3, the
validated diagnostics disagree: probes call the facts removed and relearning does not. The one cheap test
that flagged every unlearned model was the answer log-probability compared with a
never-taught reference. Claims that a method removes knowledge should be checked
against anchors of this kind before they are trusted.

\appendix

\section{Two evaluation bugs}\label{app:bugs}

An earlier version of this analysis, never published, contained two bugs. Both are
silent and both apply to other work.

\paragraph{Left padding with absolute position embeddings.} GPT-2 uses learned
absolute position embeddings. With a left-padded batch, the Hugging Face forward call
does not derive position ids from the attention mask unless they are passed, so a
padded prompt is read at shifted positions. On \texttt{M\_injected} one forget prompt
scored $\log p = -3.94$ for the correct answer in a padded batch and $-0.0012$ alone.
With the fix (position ids from the cumulative attention mask), \texttt{M\_injected}'s
retain and forget accuracy went from 0.33 and 0.38 to 1.0 and 1.0. Every behavioural
number in the earlier version, and the matched band built on them, was affected; all
checkpoints in this paper were re-selected after the fix.

\paragraph{Mismatched start-of-sequence token.} Activations had been extracted with a
beginning-of-sequence token that the models never saw in training or in behavioural
scoring. All activations in this paper are extracted without it.

\section{RMU on small models}\label{app:rmu}\label{app:deviations}

RMU failed three times on GPT-2 before the configuration reported here, and the
failures are informative. With an unmasked mean-squared steering loss at block 6,
almost all of the update went into moving the first position's residual, which in
GPT-2 is an attention sink \citep{xiao2024sinks} with norm about 3000 against about
87 elsewhere; forget accuracy stayed at 1.0 while perplexity rose. Masking padding and
position 0 fixed the sink, but the forget and retain prompts share their templates,
so steering every forget token pulled against the retain term on the same template
tokens, and forget accuracy stopped at 0.99. Restricting the steering loss to the
read-out positions (the last prompt token and the answer) and moving it to block 9,
where the city becomes linearly readable, still did not reach the band on GPT-2. The
same read-out configuration, scaled to block 21 of 28, did reach it on Qwen3 after
1000 steps. Every attempt, including those excluded, is in the released logs.

\section{Hyperparameters}\label{app:hparams}

\begin{table}[h]
\caption{Training settings}
\label{tab:hparams}
\centering
\small
\begin{tabular}{p{0.19\linewidth}p{0.35\linewidth}p{0.35\linewidth}}
\toprule
 & GPT-2 Small & Qwen3-0.6B-Base \\
\midrule
Injection & lr $2\times10^{-5}$, batch 8, 4 epochs & lr $10^{-5}$, 3 epochs (first epoch with forget and retain $\geq 0.95$) \\
Oracle & same as injection, 540 records, seeds 0, 1, 2 & same \\
NPO / GD & lr $10^{-5}$, batch 4, eval every 40 & ladder $\{1, 2, 5\}\times10^{-5}$, batch 4, eval every 20 \\
Suppressed anchors & NPO, $k=11$ (lr $10^{-4}$), $k=9$ (lr $5\times10^{-5}$) & NPO, $k=27$ (lr $10^{-4}$), $k=21$ (lr $5\times10^{-5}$) \\
RMU & block 9, MLPs of blocks 6 to 8, lr $5\times10^{-5}$ & block 21, MLPs of blocks 18 to 20, lr $2\times10^{-4}$, 1500 steps \\
Band & \multicolumn{2}{p{0.72\linewidth}}{forget $\leq 0.15$, retain $\geq 0.8\times$ injected, perplexity $\leq 1.2\times$ injected (NPO$^\dagger$ on Qwen3: $\leq 1.35\times$)} \\
\bottomrule
\end{tabular}
\end{table}

\bibliographystyle{plainnat}
\bibliography{refs}

\begin{thebibliography}{26}
\providecommand{\natexlab}[1]{#1}
\providecommand{\url}[1]{\texttt{#1}}
\expandafter\ifx\csname urlstyle\endcsname\relax
  \providecommand{\doi}[1]{doi: #1}\else
  \providecommand{\doi}{doi: \begingroup \urlstyle{rm}\Url}\fi

\bibitem[Alain and Bengio(2016)]{alain2016probes}
Guillaume Alain and Yoshua Bengio.
\newblock Understanding intermediate layers using linear classifier probes.
\newblock \emph{arXiv preprint}, 2016.
\newblock {arXiv:1610.01644}.

\bibitem[Belrose et~al.(2023{\natexlab{a}})Belrose, Furman, Smith, Halawi,
  Ostrovsky, McKinney, Biderman, and Steinhardt]{belrose2023tunedlens}
Nora Belrose, Zach Furman, Logan Smith, Danny Halawi, Igor Ostrovsky, Lev
  McKinney, Stella Biderman, and Jacob Steinhardt.
\newblock Eliciting latent predictions from transformers with the tuned lens.
\newblock \emph{arXiv preprint}, 2023{\natexlab{a}}.
\newblock {arXiv:2303.08112}.

\bibitem[Belrose et~al.(2023{\natexlab{b}})Belrose, Schneider-Joseph, Ravfogel,
  Cotterell, Raff, and Biderman]{belrose2023leace}
Nora Belrose, David Schneider-Joseph, Shauli Ravfogel, Ryan Cotterell, Edward
  Raff, and Stella Biderman.
\newblock {LEACE}: Perfect linear concept erasure in closed form.
\newblock In \emph{Advances in Neural Information Processing Systems
  (NeurIPS)}, 2023{\natexlab{b}}.
\newblock arXiv:2306.03819.

\bibitem[Bourtoule et~al.(2021)Bourtoule, Chandrasekaran, Choquette-Choo, Jia,
  Travers, Zhang, Lie, and Papernot]{bourtoule2021sisa}
Lucas Bourtoule, Varun Chandrasekaran, Christopher~A. Choquette-Choo, Hengrui
  Jia, Adelin Travers, Baiwu Zhang, David Lie, and Nicolas Papernot.
\newblock Machine unlearning.
\newblock In \emph{IEEE Symposium on Security and Privacy}, pages 141--159,
  2021.
\newblock \doi{10.1109/SP40001.2021.00019}.

\bibitem[Cao and Yang(2015)]{cao2015forget}
Yinzhi Cao and Junfeng Yang.
\newblock Towards making systems forget with machine unlearning.
\newblock In \emph{IEEE Symposium on Security and Privacy}, pages 463--480,
  2015.
\newblock \doi{10.1109/SP.2015.35}.

\bibitem[Deeb and Roger(2024)]{deeb2024remove}
Aghyad Deeb and Fabien Roger.
\newblock Do unlearning methods remove information from language model weights?
\newblock \emph{arXiv preprint}, 2024.
\newblock {arXiv:2410.08827}.

\bibitem[Dorna et~al.(2025)Dorna, Mekala, Zhao, McCallum, Lipton, Kolter, and
  Maini]{dorna2025openunlearning}
Vineeth Dorna, Anmol Mekala, Wenlong Zhao, Andrew McCallum, Zachary~C. Lipton,
  J.~Zico Kolter, and Pratyush Maini.
\newblock {OpenUnlearning}: Accelerating {LLM} unlearning via unified
  benchmarking of methods and metrics.
\newblock \emph{arXiv preprint}, 2025.
\newblock {arXiv:2506.12618}.

\bibitem[Fan et~al.(2024)Fan, Liu, Lin, Jia, Zhang, Mei, and
  Liu]{fan2024simnpo}
Chongyu Fan, Jiancheng Liu, Licong Lin, Jinghan Jia, Ruiqi Zhang, Song Mei, and
  Sijia Liu.
\newblock Simplicity prevails: Rethinking negative preference optimization for
  {LLM} unlearning.
\newblock \emph{arXiv preprint}, 2024.
\newblock {arXiv:2410.07163}.

\bibitem[Hewitt and Liang(2019)]{hewitt2019control}
John Hewitt and Percy Liang.
\newblock Designing and interpreting probes with control tasks.
\newblock In \emph{Conference on Empirical Methods in Natural Language
  Processing (EMNLP)}, pages 2733--2743, 2019.
\newblock \doi{10.18653/v1/D19-1275}.

\bibitem[Hong et~al.(2024)Hong, Yu, Yang, Ravfogel, and
  Geva]{hong2024intrinsic}
Yihuai Hong, Lei Yu, Haiqin Yang, Shauli Ravfogel, and Mor Geva.
\newblock Intrinsic test of unlearning using parametric knowledge traces.
\newblock \emph{arXiv preprint}, 2024.
\newblock {arXiv:2406.11614}.

\bibitem[Hu et~al.(2024)Hu, Fu, Wu, and Smith]{hu2024jogging}
Shengyuan Hu, Yiwei Fu, Zhiwei~Steven Wu, and Virginia Smith.
\newblock Unlearning or obfuscating? {Jogging} the memory of unlearned {LLMs}
  via benign relearning.
\newblock \emph{arXiv preprint}, 2024.
\newblock {arXiv:2406.13356}.

\bibitem[Li et~al.(2024)Li, Pan, Gopal, Yue, Berrios, Gatti, Li, Dombrowski,
  Goel, Phan, et~al.]{li2024wmdp}
Nathaniel Li, Alexander Pan, Anjali Gopal, Summer Yue, Daniel Berrios, Alice
  Gatti, Justin~D. Li, Ann-Kathrin Dombrowski, Shashwat Goel, Long Phan, et~al.
\newblock The {WMDP} benchmark: Measuring and reducing malicious use with
  unlearning.
\newblock In \emph{International Conference on Machine Learning (ICML)}, 2024.
\newblock arXiv:2403.03218.

\bibitem[Loshchilov and Hutter(2019)]{loshchilov2019adamw}
Ilya Loshchilov and Frank Hutter.
\newblock Decoupled weight decay regularization.
\newblock In \emph{International Conference on Learning Representations
  (ICLR)}, 2019.

\bibitem[{\L}ucki et~al.(2024){\L}ucki, Wei, Huang, Henderson, Tram{\`e}r, and
  Rando]{lucki2024adversarial}
Jakub {\L}ucki, Boyi Wei, Yangsibo Huang, Peter Henderson, Florian Tram{\`e}r,
  and Javier Rando.
\newblock An adversarial perspective on machine unlearning for {AI} safety.
\newblock \emph{arXiv preprint}, 2024.
\newblock {arXiv:2409.18025}.

\bibitem[Lynch et~al.(2024)Lynch, Guo, Ewart, Casper, and
  Hadfield-Menell]{lynch2024eight}
Aengus Lynch, Phillip Guo, Aidan Ewart, Stephen Casper, and Dylan
  Hadfield-Menell.
\newblock Eight methods to evaluate robust unlearning in {LLMs}.
\newblock \emph{arXiv preprint}, 2024.
\newblock {arXiv:2402.16835}.

\bibitem[Maini et~al.(2024)Maini, Feng, Schwarzschild, Lipton, and
  Kolter]{maini2024tofu}
Pratyush Maini, Zhili Feng, Avi Schwarzschild, Zachary~C. Lipton, and J.~Zico
  Kolter.
\newblock {TOFU}: A task of fictitious unlearning for {LLMs}.
\newblock In \emph{Conference on Language Modeling (COLM)}, 2024.
\newblock arXiv:2401.06121.

\bibitem[{nostalgebraist}(2020)]{nostalgebraist2020logitlens}
{nostalgebraist}.
\newblock Interpreting {GPT}: The logit lens.
\newblock LessWrong, 2020.
\newblock URL
  \url{https://www.lesswrong.com/posts/AcKRB8wDpdaN6v6ru/interpreting-gpt-the-logit-lens}.

\bibitem[Patil et~al.(2024)Patil, Hase, and Bansal]{patil2024deleted}
Vaidehi Patil, Peter Hase, and Mohit Bansal.
\newblock Can sensitive information be deleted from {LLMs}? {Objectives} for
  defending against extraction attacks.
\newblock In \emph{International Conference on Learning Representations
  (ICLR)}, 2024.
\newblock arXiv:2309.17410.

\bibitem[{Qwen Team}(2025)]{qwen2025qwen3}
{Qwen Team}.
\newblock Qwen3 technical report.
\newblock \emph{arXiv preprint}, 2025.
\newblock {arXiv:2505.09388}.

\bibitem[Radford et~al.(2019)Radford, Wu, Child, Luan, Amodei, and
  Sutskever]{radford2019gpt2}
Alec Radford, Jeffrey Wu, Rewon Child, David Luan, Dario Amodei, and Ilya
  Sutskever.
\newblock Language models are unsupervised multitask learners.
\newblock Technical report, OpenAI, 2019.

\bibitem[Shi et~al.(2024)Shi, Lee, Huang, Malladi, Zhao, Holtzman, Liu,
  Zettlemoyer, Smith, and Zhang]{shi2024muse}
Weijia Shi, Jaechan Lee, Yangsibo Huang, Sadhika Malladi, Jieyu Zhao, Ari
  Holtzman, Daogao Liu, Luke Zettlemoyer, Noah~A. Smith, and Chiyuan Zhang.
\newblock {MUSE}: Machine unlearning six-way evaluation for language models.
\newblock \emph{arXiv preprint}, 2024.
\newblock {arXiv:2407.06460}.

\bibitem[Xiao et~al.(2024)Xiao, Tian, Chen, Han, and Lewis]{xiao2024sinks}
Guangxuan Xiao, Yuandong Tian, Beidi Chen, Song Han, and Mike Lewis.
\newblock Efficient streaming language models with attention sinks.
\newblock In \emph{International Conference on Learning Representations
  (ICLR)}, 2024.
\newblock arXiv:2309.17453.

\bibitem[Xu et~al.(2025)Xu, Yue, Liu, Ye, Hu, and Du]{xu2025isntdeletion}
Xiaoyu Xu, Xiang Yue, Yang Liu, Qingqing Ye, Haibo Hu, and Minxin Du.
\newblock Unlearning isn't deletion: Investigating reversibility of machine
  unlearning in {LLMs}.
\newblock \emph{arXiv preprint}, 2025.
\newblock {arXiv:2505.16831}.

\bibitem[Yang and Yeung(2026)]{yang2026restoration}
Sen Yang and Yuen-Hei Yeung.
\newblock Unlearning as distribution restoration: A controlled counterfactual
  study, a validated selective screen, and the limits of oracle-free
  certification.
\newblock \emph{arXiv preprint}, 2026.
\newblock {arXiv:2607.19442}.

\bibitem[Yao et~al.(2023)Yao, Xu, and Liu]{yao2023llmunlearning}
Yuanshun Yao, Xiaojun Xu, and Yang Liu.
\newblock Large language model unlearning.
\newblock \emph{arXiv preprint}, 2023.
\newblock {arXiv:2310.10683}.

\bibitem[Zhang et~al.(2024)Zhang, Lin, Bai, and Mei]{zhang2024npo}
Ruiqi Zhang, Licong Lin, Yu~Bai, and Song Mei.
\newblock Negative preference optimization: From catastrophic collapse to
  effective unlearning.
\newblock In \emph{Conference on Language Modeling (COLM)}, 2024.
\newblock arXiv:2404.05868.

\end{thebibliography}

\end{document}